\pdfoutput=1

\documentclass[a4paper,10pt,conference]{ieeeconf}

\IEEEoverridecommandlockouts

\usepackage{graphicx} 
\usepackage{tabularx} 
\usepackage{multirow} 
\usepackage{makecell} 
\usepackage{balance} 
\usepackage[hyphens]{url} 

\newcolumntype{C}{>{\centering\arraybackslash}X}
\newcolumntype{L}{>{\raggedright\arraybackslash}X}
\newcolumntype{R}{>{\raggedleft\arraybackslash}X}

\title{\LARGE\bf
    Comparative Evaluation of an XR Pen-based Control Interface for\\
    Semi-Autonomous Mobile Robot Navigation in Service Environments
}

\author{
    Alicia Torck$^{1}$,
    Carl Tornberg$^{2}$,
    Eric Piette$^{1}$,
    Renaud Ronsse$^{1}$,
    Benoit Macq$^{1}$,\\
    Gustavo Alfonso Garcia Ricardez$^{2}$,
    Lotfi El Hafi$^{2, *}$,
    and Tadahiro Taniguchi$^{2, 3}$
    \thanks{
        This work was supported by the Japan Science and Technology Agency~(JST), Moonshot Research~\& Development Program, Grant Number JPMJMS2011, and by the Japan Society for the Promotion of Science~(JSPS), KAKENHI Grants-in-Aid for Scientific Research, Grant Number JP22K17981.
    }
    \thanks{
        $^{1}$Alicia Torck, Eric Piette, Renaud Ronsse, and Benoit Macq are with Université catholique de Louvain (UCLouvain);
        1 Place de l'Université, Louvain-la-Neuve 1348, Belgium. 
        Eric Piette and Benoit Macq are with its Institute of Information and Communication Technologies, Electronics and Applied Mathematics~(ICTEAM), Renaud Ronsse with its Institute of Mechanics, Materials, and Civil Engineering~(iMMC), and Alicia Torck with both.
        {\tt\small alicia.torck @student.uclouvain.be, \{eric.piette, renaud.ronsse, benoit.macq\}@uclouvain.be}
    }
    \thanks{
        $^{2}$Carl Tornberg, Gustavo Alfonso Garcia Ricardez, Lotfi El Hafi, and Tadahiro Taniguchi are with Ritsumeikan University;
        1-1-1 Noji-Higashi, Kusatsu, Shiga 525-8577, Japan. 
        Carl Tornberg is with its Graduate School of Information Science and Engineering, and the others with its Research Organization of Science and Technology.
        {\tt\small \{tornberg.carl, garcia-g, lotfi.elhafi, taniguchi\}@em.ci.ritsumei.ac.jp}
    }
    \thanks{
        $^{3}$Tadahiro Taniguchi is with the Graduate School of Informatics of Kyoto University;
        Yoshida-Honmachi, Sakyo, Kyoto 606-8501, Japan. 
        {\tt\small taniguchi@i.kyoto-u.ac.jp}
    }
    \thanks{
        $^{*}$Corresponding author.
    }
}

\begin{document}


\maketitle
\thispagestyle{empty}
\pagestyle{empty}


\begin{abstract}
    Service robots remain difficult to deploy in domestic environments, partly because fully autonomous operation is not yet reliable in unpredictable surroundings, and partly because conventional control methods remain inaccessible to novice users.
    Extended Reality~(XR) enables operators to visualize robot information overlaid onto the real world and to interact with augmented elements. Yet, common XR control methods, such as motion controllers and hand gestures, are still perceived as unintuitive.
    This paper presents a control interface that uses a commercial XR pen to command a semi-autonomous mobile robot in Augmented Reality~(AR): the operator points at a position in the room, selects it, and drags an augmented arrow to set the desired orientation of the robot at this destination.
    Two additional interfaces, based on the XR motion controllers and hand gestures, were developed within the same framework.
    To assess the performance and users' perception of these interfaces, and of the XR pen in particular, a study with 10 participants compared four control methods, i.e., the XR pen, the XR motion controllers, hand gestures, and a computer-based baseline RViz, in navigation tasks performed in a home-like environment.
    Repeated-measures analyses show that the control method significantly affects the perceived workload and usability: the XR motion controllers obtain the best mean scores, comparable to those of the computer-based baseline, whereas hand gestures lead to a significantly higher workload than both. The XR pen yields the shortest mean task selection time with the smallest variability, but this difference does not reach statistical significance, and a tracking issue of the XR pen, caused by the headset switching to hand tracking, held back its user experience.
\end{abstract}


\begin{figure}[t]
    \centering
    \includegraphics[width=0.94\linewidth]{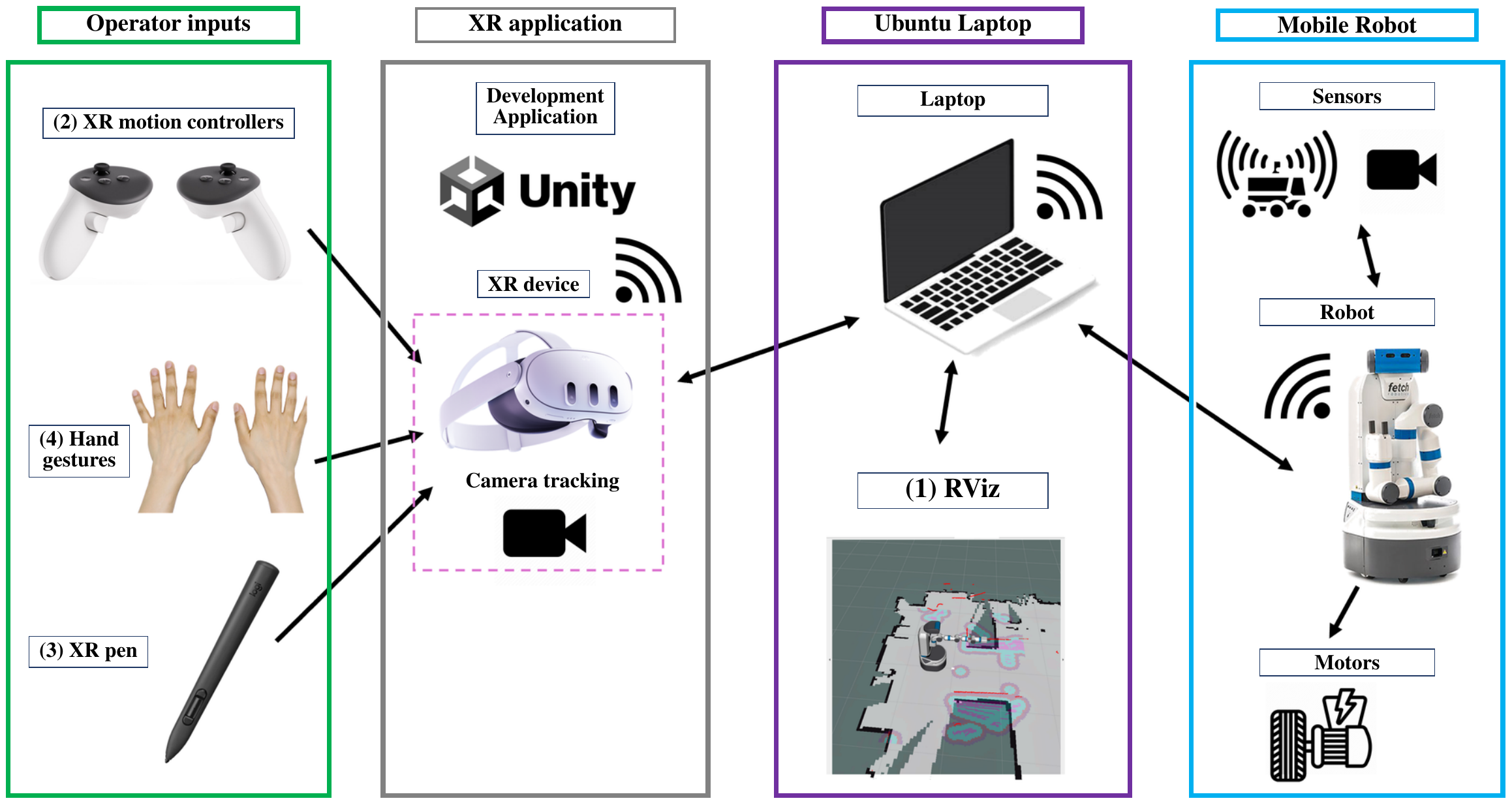}
    \caption{
        System overview of the proposed approach.
        Each colored box groups the components running on one device, and the four control methods are numbered (1) to (4) throughout the paper.
        The XR inputs (2)--(4) are tracked by the headset running the Unity-based XR application, which sends the selected navigation goals to a semi-autonomous mobile robot over the network, while (1)~the computer-based RViz baseline runs on the laptop.
    }
    \label{fig:approach_diagram}
\end{figure}


\section{Introduction}
\label{sec:introduction}

Although robots have become efficient tools in industrial, medical, and service environments~\cite{gonzalez-aguirre_service_2021, garcia_ricardez_autonomous_2022}, their deployment in domestic environments remains mostly limited to vacuum cleaners and lawn mowers~\cite{zachiotis_survey_2018}, slowed down by their cost, lack of transparency of their decisions, and unreliability in unpredictable environments.
Non-expert users often struggle to predict the intentions of such robots and do not feel comfortable around them~\cite{suzuki_augmented_2022, kennel-maushart_interacting_2023}.

Human-Robot Collaboration~(HRC) through semi-autonomous control provides a middle ground by combining robot and human reasoning, allowing the user to remain in control of the robot and gain confidence in it.
This solution is already deployed in factories with collaborative robots, but the associated control and programming methods require training that is not suitable for novice users~\cite{palmarini_designing_2018}.
Extended Reality~(XR) presents a promising framework to address this accessibility gap:
using an XR headset, the operator can visualize overlaid robot information, such as future paths and waypoints, directly in the real-world environment~\cite{walker_communicating_2018, cleaver_enhancing_2024}, without splitting attention between the scene and a separate screen.

XR also introduces new control methods: motion controllers offer buttons and joysticks to manipulate augmented elements or control the motion of a robot~\cite{nakanishi_towards_2020, luo_user-customizable_2024}, while hand tracking maps detected hand gestures, such as pinch, to control inputs~\cite{tornberg_mixed_2024}.
Selection-based control methods built on these devices allow users to point at and select targets for manipulators~\cite{tadeja_using_2024} or mobile robots~\cite{kennel-maushart_interacting_2023}.
However, such methods remain unintuitive for non-expert users:
the pinch gesture used to trigger selections has repeatedly been reported as a source of difficulty and frustration~\cite{tadeja_using_2024, kennel-maushart_interacting_2023}.
Furthermore, comparative studies of XR selection devices that evaluate perceived workload, fatigue, and usability in scenarios of various difficulties are needed to refine our understanding of this usability gap~\cite{yu_object_2024}.
Pen-like devices have shown great potential to improve pointing accuracy and comfort~\cite{pham_pen_2019, ong_augmented_2020}:
held like a familiar pen, a tracked stylus offers higher precision than motion controllers~\cite{waugh_could_2026} and, being lighter, is expected to reduce the physical demand.
However, they have rarely been included in comparative studies for robot control in XR, notably because both of these studies relied on custom-made prototypes.

The problem addressed in this study is thus twofold.
First, conventional and XR-based control methods remain insufficiently intuitive for novice users.
Second, an XR pen, despite its potential for accurate pointing, has not been objectively compared with established control methods for mobile robot navigation.
This study therefore addresses two research questions:
\begin{itemize}
    \item Q1) Will users rate an XR pen as an intuitive control method that is easy to use and learn?
    \item Q2) Will an XR pen improve task efficiency by yielding shorter task completion, selection, and waiting times?
\end{itemize}
From prior findings in the literature, four hypotheses are formulated, two of them concerning the user's perception of the control interfaces:
\begin{itemize}
    \item H1) XR-based control interfaces improve the user experience and decrease the user's workload compared to a computer-based control interface.
    \item H2) A control interface based on an XR pen enhances the user experience and decreases the user's workload compared to other XR-based control interfaces.
\end{itemize}
The remaining two concern the objective task efficiency:
\begin{itemize}
    \item H3) XR-based control interfaces enhance the navigation task efficiency compared to a computer-based control interface by reducing the task selection times, but not for difficult and occluded target locations.
    \item H4) A control interface based on an XR pen improves the navigation task efficiency by yielding shorter task selection times compared to other XR-based control interfaces.
\end{itemize}
Hand gestures are expected to frustrate users through the pinch gesture~\cite{tadeja_using_2024, kennel-maushart_interacting_2023}, the XR pen to reduce the physical demand and selection times~\cite{pham_pen_2019}, and the computer-based map to remain advantageous for occluded targets.

To address these questions, this paper presents a novel interface using a commercial XR pen to control a semi-autonomous mobile robot in Augmented Reality~(AR) by selecting navigation goals, as shown in Fig.~\ref{fig:approach_diagram}.
The operator points at a position in the room, selects it, and drags an augmented arrow to set the desired orientation of the robot at its destination.
Two additional interfaces were developed within the same framework for two other XR control methods: the XR motion controllers and the operator's hand gestures, resulting in a four-way comparison with the computer-based baseline RViz.
The contributions of this study are therefore twofold:
\begin{enumerate}
    \item The integration of a navigation control interface based on a commercial XR pen alongside two interfaces based on the XR motion controllers and hand gestures, within the same AR framework connected to a semi-autonomous mobile robot.
    \item A comparative evaluation of the four control interfaces in a user study with 10 participants, combining standardized workload and usability questionnaires with objective navigation metrics.
\end{enumerate}


\begin{figure*}[t]
    \centering
    \includegraphics[width=0.9\textwidth]{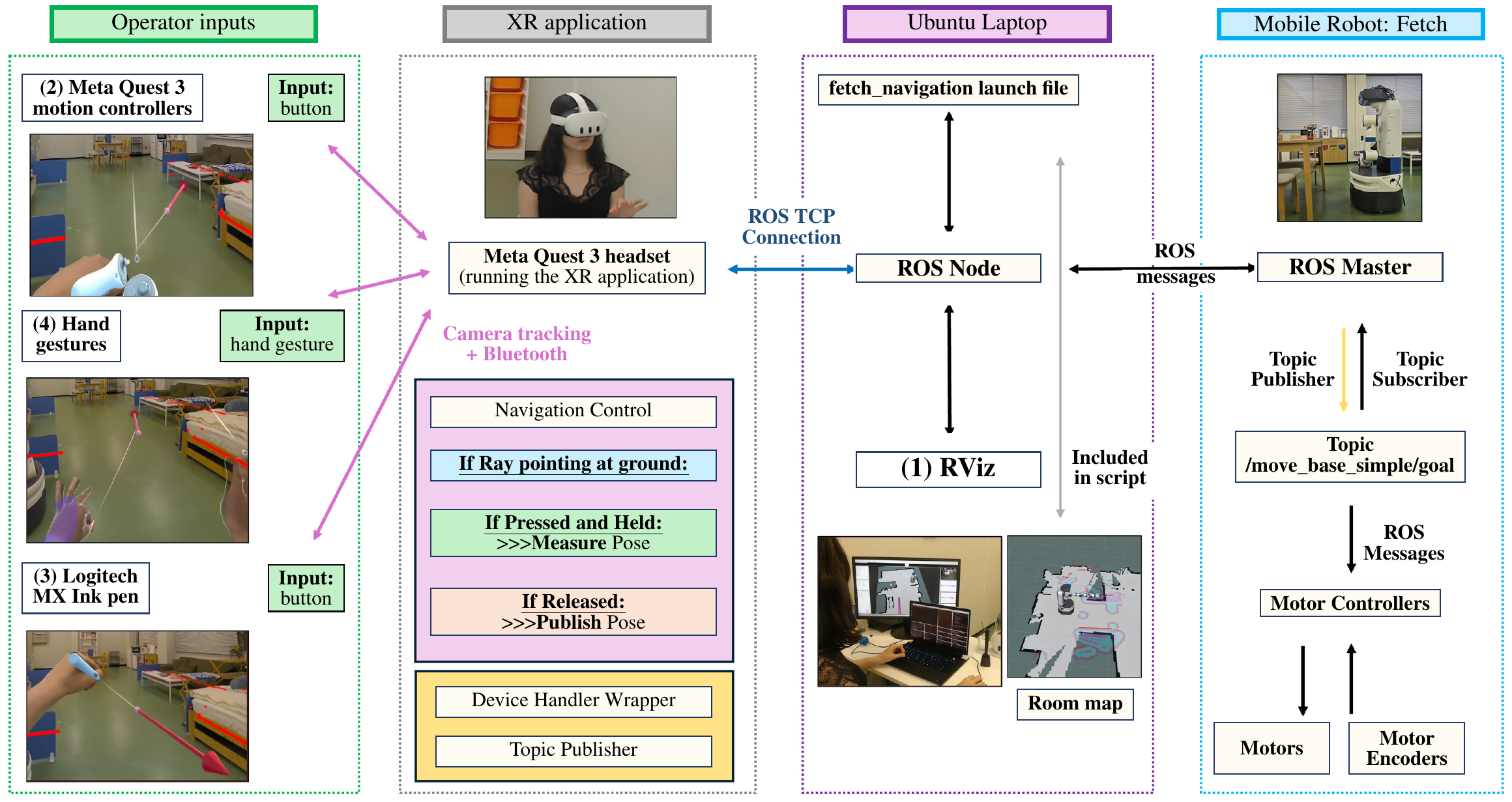}
    \caption{
        Integration of the proposed approach, numbered as in Fig.~\ref{fig:approach_diagram}: the scripts of the XR application onboard the headset detect the selection inputs of the tracked devices, shown with the corresponding operator's point of view, and publish the navigation goals through a ROS TCP connection to a node on the laptop, which relays them to the navigation topic of the robot and also runs (1)~the RViz baseline.
    }
    \label{fig:implementation_diagram}
\end{figure*}


\section{Related Work}
\label{sec:related_work}


\subsection{Extended Reality for Human-Robot Interaction}

On the reality-virtuality continuum~\cite{milgram_augmented_1995}, AR describes systems that augment physical objects or surrounding environments in the real world, while XR is the umbrella term for AR, Augmented Virtuality~(AV), and VR, whose impact on HRI was evaluated in a follow-up of this study~\cite{tornberg_evaluating_2026}.
In this paper, XR names the devices and interfaces of the headset, which all operate in AR.
A complete taxonomy of AR technologies for HRI is provided in~\cite{suzuki_augmented_2022}, organized by visual effects, hardware placements, and interaction modalities, among which head-mounted displays are the most popular in service robotics, as they require no installation and suit any environment.
Mixed Reality~(MR) headsets have notably been used to let non-expert users teach service robots and visualize their perception during object categorization~\cite{el_hafi_teaching_2021, nakamura_multimodal_2022}, as well as to convey robot motion intent and safety indices to human coworkers~\cite{garcia_ricardez_toward_2023}.
In~\cite{walker_communicating_2018}, four AR visual cues communicating the motion intent of an aerial robot were compared, demonstrating that displaying its future path or waypoints significantly improves task efficiency and the perception of the robot as a work companion.
In~\cite{ong_augmented_2020}, the lack of intuitiveness of conventional industrial robot programming was addressed by combining an AR headset with a wireless handheld pointer to define paths in the workspace of a manipulator, and novice participants rated the interface as user-friendly and intuitive.


\subsection{XR-based Robot Control Methods}

In~\cite{tadeja_using_2024}, eye-gaze and hand-ray pointing interfaces were compared to designate the objects to be picked by a collaborative manipulator during an assembly task.
The eye-gaze interface led to shorter completion and selection times and a smoother user experience, whereas the hand-ray interface, triggered by a pinch gesture, was less favored.
In~\cite{kennel-maushart_interacting_2023}, hand tracking with an MR headset was compared against touch gestures on a tablet to command a Multi-Robot System~(MRS).
Users preferred the tablet for simpler tasks but performed better with hand tracking for more complex ones, while the pinch gesture was frequently mentioned as unintuitive and a source of frustration.
Brain-computer interfaces combined with eye tracking have also been proposed to select the formations of a robot fleet in AR~\cite{xu_combination_2024}, although eye tracking is not available in several recently released headsets, such as the Meta Quest 3 used in this study.
In this study, we adopt the same selection-based control paradigm as~\cite{tadeja_using_2024, kennel-maushart_interacting_2023}, but we compare three XR control methods available on a recent consumer headset, including the rarely evaluated XR pen, against RViz, the 3D visualizer for the Robot Operating System~(ROS)~\cite{quigley_ros_2009} commonly used in mobile robotics, as computer-based baseline.


\subsection{Pen-Like Control Devices}

Object selection and manipulation with XR headsets are reviewed in~\cite{yu_object_2024}, identifying research challenges directly related to this study.
These include the need for selection techniques suited to complex scenarios with distant or occluded targets, the lack of comparative usability studies between devices, and ergonomic concerns about workload and fatigue, while ray-casting techniques based on the user's hands, initially considered intuitive, proved imprecise in difficult situations.
In~\cite{pham_pen_2019}, a custom pen-like pointing device was compared with motion controllers and a 2D mouse in a target selection task performed in both Virtual Reality~(VR) and AR.
The pen was comparable to the mouse, outperformed the motion controllers, and was preferred in terms of comfort.
However, the results of such studies are difficult to compare objectively, as each of them relied on its own custom-made prototype, a limitation that commercial 6-Degree-of-Freedom~(DOF) styluses for consumer headsets now help overcome~\cite{kehoe_logitech_2025}, also enabling studies of their specific pointing behaviors, which showed them less affected than controllers by the Heisenberg effect, in which pressing a button perturbs the pointing pose~\cite{waugh_could_2026}.
Previous works further recommend conducting user studies with larger robots and scenarios including occluded targets that force the user to move~\cite{kennel-maushart_interacting_2023, yu_object_2024}.
We address these recommendations with a commercial XR pen and a human-sized mobile robot in a home-like environment with occluded targets, assessing the perceived workload and usability alongside navigation metrics.


\section{Proposed Approach}
\label{sec:proposed_approach}

This study proposes a control method based on navigation target selection for semi-autonomous mobile robots in AR, using an XR pen, evaluated through the comparative study defined by Hypotheses H1 to H4.
The robot is controlled in navigation only:
the operator selects the position and orientation that the robot must reach, and the robot navigates autonomously to this pose.
The XR-based control method reproduces the framework of RViz, designed for intuitiveness and control accuracy.
Wearing an XR headset, the operator points at a target position in the room using a ray cast from the connected input and triggers a selection action.
A virtual target and an augmented arrow appear at the selected position:
while holding the selection input, the operator drags the arrow to set the orientation the robot should face at its destination.
When the input is released, the complete pose is sent to the robot, which navigates to it autonomously.
The same technique is used for all three XR devices, and only the selection input varies, whereas with the computer-based baseline the operator selects a position on a map of the room displayed in RViz and drags the orientation arrow using the mouse.
Besides the RViz baseline, the motion controllers and hand gestures were chosen as the built-in inputs of the headset.
The comparison between the XR interfaces thus isolates the input device, whereas the comparison with RViz changes both the input device and the visualization, whose respective effects cannot be separated.
The headset tracks the XR input, displays the virtual elements, and sends the completed pose to the robot, while a laptop relays the connection and runs RViz (Fig.~\ref{fig:approach_diagram}).


\section{System Integration}
\label{sec:system_integration}


\subsection{Hardware Components}

The mobile robot is a Fetch Mobile Manipulator~\cite{wise_fetch_2016}, a human-sized robot operated with ROS, composed of a mobile base, a 7-DOF arm, a gripper, and a 2-DOF head, of which only the semi-autonomous navigation of the base is used.
The XR headset is a Meta Quest 3\footnote{\url{https://www.meta.com/quest/quest-3/}}, a standalone headset working in passthrough mode, selected for its large field of view and graphical capabilities, although it does not provide eye tracking.
The XR pen is a Logitech MX Ink~\cite{kehoe_logitech_2025}, a 6-DOF pen-like control device developed for the Meta Quest headsets, composed of four buttons and a pressure-sensitive tip, and designed to provide high precision and low latency while being lighter than the XR motion controllers.


\subsection{Software Integration}

The XR application managing the visual effects and control interactions was developed with Unity and runs onboard the headset, where it was integrated into an existing XR project for robot control developed in previous work~\cite{tornberg_mixed_2024}, following the integration shown in Fig.~\ref{fig:implementation_diagram}.
Three main scripts implement the proposed control interfaces.
A topic publisher sends ROS messages from the XR application to the robot through the network.
A device handler manages the connections and disconnections of the different XR devices, enabling the integration of the XR pen alongside the other XR controllers without input conflicts.
A navigation control script casts a ray from the connected input towards the ground and detects the control inputs: the 2D intersection between the ray and the ground is saved with reference to the robot's frame, the orientation of the augmented arrow is updated as a quaternion while the input is held, and the resulting pose is published on release as a navigation goal message to the \texttt{/move\_base\_simple/goal} topic controlling the autonomous navigation of the robot.
The application connects to the ROS network through a TCP connector\footnote{\url{https://github.com/Unity-Technologies/ROS-TCP-Connector}} to a node running on an Ubuntu laptop, while the robot acts as the ROS master.
For the baseline, RViz runs on the laptop and publishes the pose selected on the map, after pressing a dedicated button, on the same ROS topic.
The ROS components were deployed in a containerized software development environment for robotics~\cite{el_hafi_abstraction-rich_2018, el_hafi_software_2022}.


\subsection{XR-based Control Interfaces}

\begin{figure}[t]
    \centering
    \frame{\includegraphics[width=0.315\linewidth]{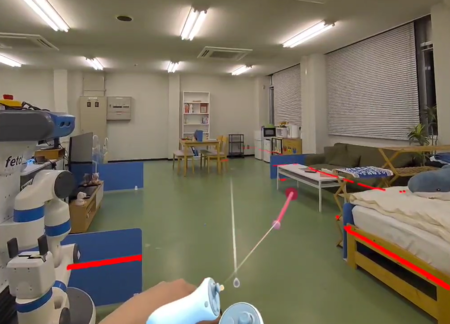}}\hfill
    \frame{\includegraphics[width=0.315\linewidth]{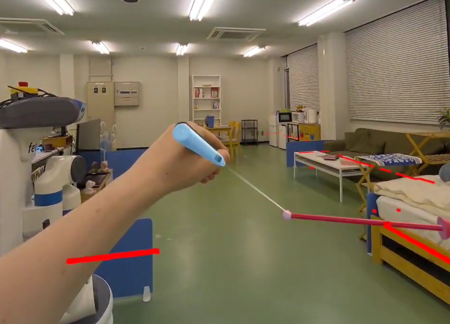}}\hfill
    \frame{\includegraphics[width=0.315\linewidth]{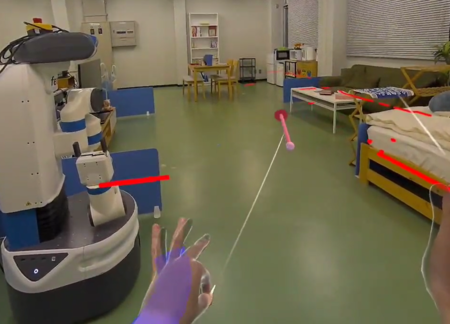}}
    \caption{
        Operator's point of view through the XR headset for (2)~the XR motion controllers, (3)~the XR pen, and (4)~hand gestures, from left to right, with the ray and the augmented arrow of the selected navigation goal.
    }
    \label{fig:interfaces_pov}
\end{figure}

The operator's point of view during a selection, with the augmented elements, is shown for each of the three XR-based interfaces in Fig.~\ref{fig:interfaces_pov}, and their control devices appear among the four compared control methods in Fig.~\ref{fig:experiment_setup} (bottom).
The XR motion controllers are two control devices, one held in each hand, and their interface uses the primary button of either controller as the selection input.
The XR pen is a single control device, and its interface uses one of the pen's side buttons.
Hand gestures use no control device:
the interface detects a pinch gesture, performed when the thumb and index fingers enter in contact, on either hand.
Only one input can trigger a selection at a time.


\section{Experiments}
\label{sec:experiments}


\subsection{Participants}

The user study, detailed in~\cite{torck_control_2025}, was conducted with 10 participants (two female, eight male) aged between 22 and 61 years old (mean: 33, standard deviation: 14), coming from six different countries.
All participants were right-handed, without encumbered hand movement or neurological condition, and eight of them had corrected vision.
Each participant first completed a demographic questionnaire including questions about prior experience with robots and XR, partly adapted from~\cite{ortenzi_robot_2022} and answered on a 5-point Likert scale~\cite{likert_technique_1932}.
The group included both domain experts and novice users: the participants reported a mean familiarity with robots of 4.2 out of 5, but little or neutral familiarity with the XR modalities.
The experiments were conducted under the Ritsumeikan University Research Ethics Guidelines, established by an independent committee, whose checklist cleared the study without the need for a formal ethics review, and the questionnaire answers were collected anonymously.


\subsection{Task and Environment}

\begin{figure}[t]
    \centering
    \includegraphics[width=0.94\linewidth]{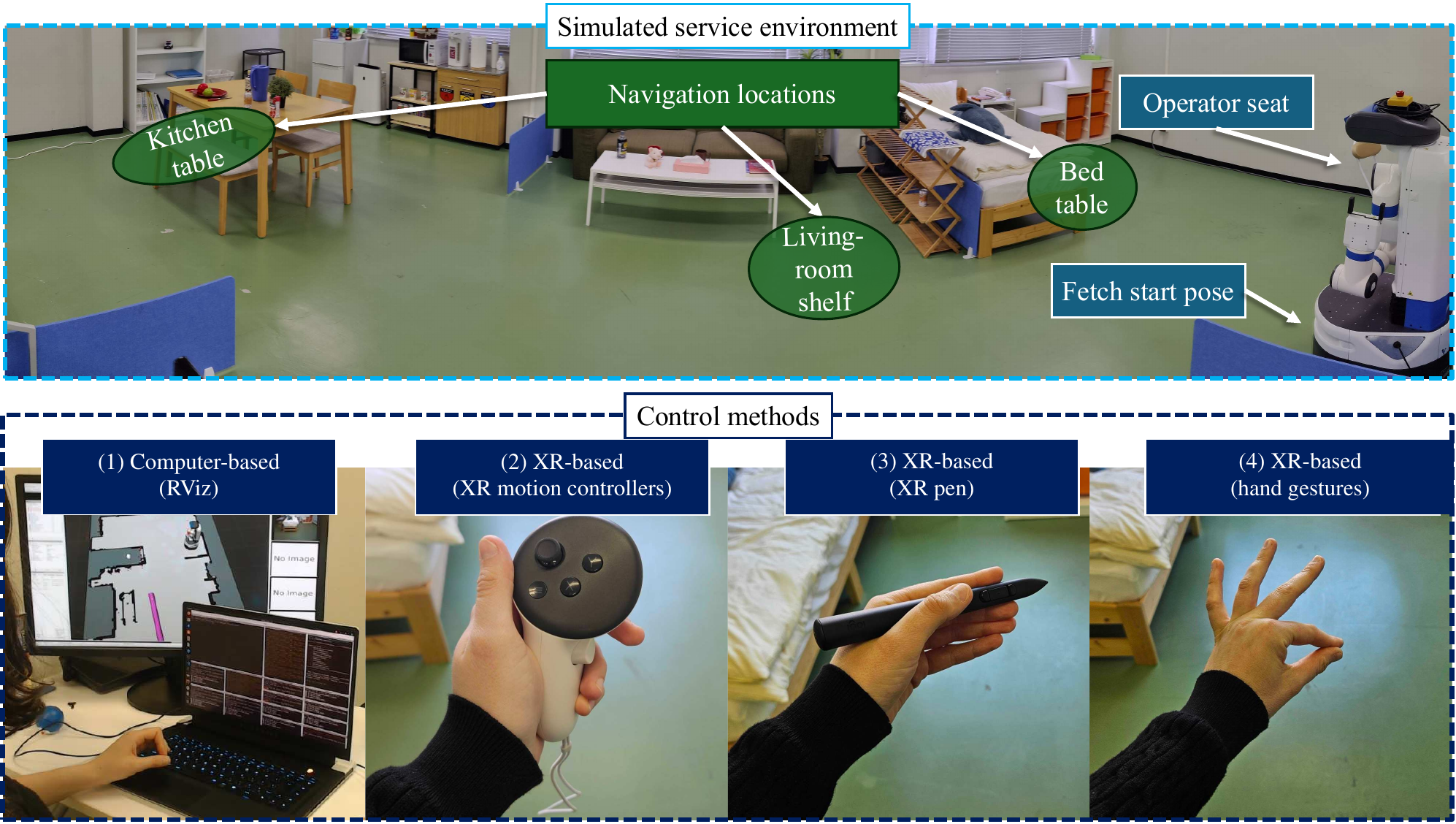}
    \caption{
        Experimental setup.
        Top: the simulated daily living environment with the three target locations, the operator seat, and the initial pose of the robot.
        Bottom: the four compared control methods, namely (1)~the computer-based RViz baseline together with the XR-based (2)~XR motion controllers, (3)~XR pen, and (4)~hand gestures.
    }
    \label{fig:experiment_setup}
\end{figure}

The experiments reproduce a service robotics scenario in which the robot must transport an object that cannot be handled by the user, such as a cup of coffee that is too hot, to a user who is unable to move from their seat.
Participants had to guide the robot to a given piece of furniture on which the object was placed, in a reproduction of a daily living environment shown in Fig.~\ref{fig:experiment_setup} (top).
The task was completed once the robot had stopped at a position close enough to the target location to grasp the object with its arm, facing it, although the arm manipulation itself was not included in this experiment.
To introduce different levels of difficulty, three target locations were used, one in each room of the home-like environment:
the kitchen table, the living room shelf, and the bed table.
The kitchen table was considered the most difficult location, as the robot had to navigate around a table and place itself in a relatively narrow space.


\subsection{Procedure}

\begin{figure}[t]
    \centering
    \includegraphics[width=0.75\linewidth]{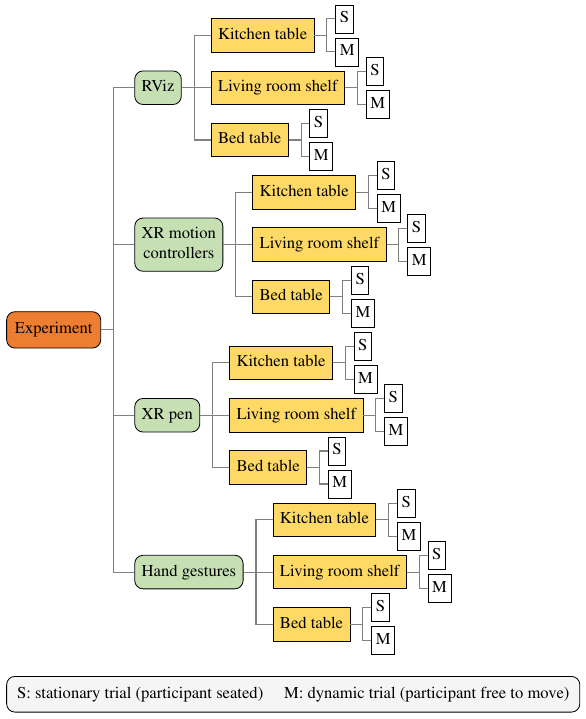}
    \caption{
        Overview of all the trial types performed by each participant: for each of the four control methods (1) to (4), one stationary trial (S) and one dynamic trial (M) for each of the three target locations, with the NASA-TLX and SUS questionnaires completed after each control method.
    }
    \label{fig:trial_types}
\end{figure}

Participants were first shown a 10-minute tutorial video describing the context of the experiment, the control methods involved, and the trial procedure, followed by a 20-minute practice session during which they could freely test the four control interfaces with the robot.
Each participant then used the four interfaces successively, performing for each interface one stationary trial, seated at the operator desk, and one dynamic trial, free to move in the room, for each of the three target locations, as illustrated in Fig.~\ref{fig:trial_types}.
The trial order was randomized per participant, first over the four interfaces, then over the three target locations within the six consecutive trials of each interface, and finally over the stationary and dynamic trials, which limits learning and order effects.
The resulting 24 trials per participant were performed only once each, as the complete experiment already lasted around two hours.
Each trial started with the robot at its initial position next to the operator desk, and ended when the participant, satisfied with the final pose of the robot, said so.
The participants received no guidance during the trials:
in particular, when the robot got stuck near obstacles due to its autonomous path planning, they were free to either wait for the robot to recover or select new navigation targets.
After testing each control method, the participant completed a NASA Task Load Index~(NASA-TLX)~\cite{hart_development_1988} questionnaire and a System Usability Scale~(SUS)~\cite{brooke_sus_1996} questionnaire, two conventional instruments in comparative studies of XR control methods~\cite{tadeja_using_2024, kennel-maushart_interacting_2023}.
At the end of the experiment, four questions on the overall interaction with the robot, based on~\cite{newbury_visualizing_2022}, were answered on a 5-point Likert scale.


\subsection{Measurements}
\label{sec:measurements}

During each trial, all navigation goal requests and their time stamps were automatically recorded, from which three duration metrics were extracted.
The task completion time spans from the start signal to the notification by the participant, as in~\cite{tadeja_using_2024}.
The task selection time, from the start of the task to the first navigation goal selection, isolates a pure interaction time not yet affected by the autonomous navigation.
The task waiting time, from the last navigation goal request to the completion of the task, measures the time spent waiting for the autonomous system.
In addition, the number of navigation requests per trial was recorded to identify the participants' navigation strategies.
The NASA-TLX questionnaires were unweighted, the raw workload index being the average of the six dimensions scaled to 100.
The questionnaire scores were analyzed with one-way repeated-measures analyses of variance~(ANOVA) over the control methods, and the duration metrics and the number of navigation requests, averaged per participant over the stationary and dynamic trials, with two-way repeated-measures ANOVAs over the control methods and the target locations.
Two missing dynamic trials of one participant were replaced by their stationary counterparts.
Violations of sphericity were handled with the Greenhouse-Geisser correction, effect sizes are reported as partial eta squared~($\eta_p^2$), and pairwise comparisons used paired $t$-tests with the Holm correction, while further paired $t$-tests compared RViz with the mean of the three XR-based interfaces.


\begin{table*}[t]
    \caption{
        Qualitative (NASA-TLX and SUS) and quantitative (task efficiency) results for the four compared control methods, numbered as in Fig.~\ref{fig:approach_diagram}, as means $\pm$ standard deviations, with the $p$-value and the effect size~$\eta_p^2$ of the control method in the repeated-measures ANOVAs described in Section~\ref{sec:measurements}.
        In each row, the best mean is in bold and the worst underlined, a lower value being better for every metric except the SUS score.
        $p$-values below the $.05$ significance level are in bold.
    }
    \begin{center}
        \small
        \setlength{\tabcolsep}{6pt}
        \renewcommand{\arraystretch}{1.0}
        \begin{tabularx}{\textwidth}{|l|CCCC|c|c|}
            \hline
            \multicolumn{1}{|c|}{Metric} & \multicolumn{1}{c}{\makecell{(1)~RViz\\(baseline)}} & \multicolumn{1}{c}{\makecell{(2)~XR motion\\controllers}} & \multicolumn{1}{c}{\makecell{(3)~XR\\pen}} & \multicolumn{1}{c|}{\makecell{(4)~Hand\\gestures}} & \multicolumn{1}{c|}{$p$-value} & \multicolumn{1}{c|}{$\eta_p^2$} \\
            \hline
            \hline
            \multicolumn{7}{|c|}{Qualitative results: user experience} \\
            \hline
            \hline
            \multicolumn{7}{|c|}{NASA Task Load Index (per dimension 0--20, raw workload 0--100)} \\
            \hline
            Mental demand & 5.65 $\pm$ 5.15 & \textbf{3.35} $\pm$ 2.83 & 7.00 $\pm$ 5.35 & \underline{7.65} $\pm$ 5.93 & .113 & .20 \\
            Physical demand & 2.65 $\pm$ 3.46 & \textbf{2.55} $\pm$ 1.38 & \underline{5.80} $\pm$ 5.31 & 3.95 $\pm$ 2.41 & .054 & .24 \\
            Temporal demand & 4.90 $\pm$ 4.43 & \textbf{4.45} $\pm$ 3.10 & 6.30 $\pm$ 4.72 & \underline{7.10} $\pm$ 4.93 & \textbf{.013} & .32 \\
            Performance & \textbf{7.55} $\pm$ 6.39 & 8.45 $\pm$ 6.05 & \underline{10.40} $\pm$ 5.66 & 9.65 $\pm$ 5.29 & .248 & .14 \\
            Effort & \textbf{4.85} $\pm$ 3.27 & 5.60 $\pm$ 4.72 & 8.10 $\pm$ 5.69 & \underline{9.35} $\pm$ 5.48 & \textbf{.030} & .28 \\
            Frustration & \textbf{5.10} $\pm$ 4.82 & 6.25 $\pm$ 5.21 & 9.90 $\pm$ 5.82 & \underline{10.60} $\pm$ 6.70 & \textbf{.004} & .39 \\
            Raw workload & 25.58 $\pm$ 18.22 & \textbf{25.54} $\pm$ 13.09 & 39.58 $\pm$ 22.10 & \underline{40.25} $\pm$ 19.33 & \textbf{.002} & .42 \\
            \hline
            \hline
            \multicolumn{7}{|c|}{System Usability Scale (0--100)} \\
            \hline
            SUS score & 68.50 $\pm$ 16.47 & \textbf{77.00} $\pm$ 12.35 & \underline{56.25} $\pm$ 26.85 & 57.50 $\pm$ 26.59 & \textbf{.043} & .26 \\
            \hline
            \hline
            \multicolumn{7}{|c|}{Quantitative results: task efficiency} \\
            \hline
            \hline
            Completion time (s) & 59.64 $\pm$ 35.84 & 62.79 $\pm$ 49.54 & \textbf{56.07} $\pm$ 37.00 & \underline{63.90} $\pm$ 41.33 & .601 & .05 \\
            Selection time (s) & 4.13 $\pm$ 5.48 & 3.06 $\pm$ 3.26 & \textbf{2.44} $\pm$ 2.56 & \underline{4.97} $\pm$ 5.16 & .105 & .20 \\
            Waiting time (s) & \underline{16.23} $\pm$ 13.74 & 15.65 $\pm$ 18.13 & 12.90 $\pm$ 12.19 & \textbf{12.56} $\pm$ 16.24 & .538 & .08 \\
            Navigation requests & \textbf{4.61} $\pm$ 4.11 & 8.38 $\pm$ 8.75 & 9.54 $\pm$ 8.46 & \underline{11.02} $\pm$ 16.95 & .064 & .23 \\
            \hline
        \end{tabularx}
    \end{center}
    \label{tab:device_results}
\end{table*}


\section{Results}
\label{sec:results}


\subsection{Qualitative Results: User Experience}

Table~\ref{tab:device_results} (qualitative results) summarizes the NASA-TLX and SUS questionnaires.
The control method had a significant effect on the raw workload ($p = .002$, $\eta_p^2 = .42$):
hand gestures led to a significantly higher workload than RViz ($p = .044$) and the motion controllers ($p = .027$), which obtained the lowest means, while the difference between the pen and the motion controllers remained just above the significance level ($p = .051$).
The temporal demand, the effort, and the frustration also depended significantly on the control method, although without significant pairwise differences.
The SUS scores also differed between control methods ($p = .043$), again without significant pairwise differences:
the motion controllers reached the highest average usability score (77), in the acceptable range, followed by RViz (68.5) in the high marginal acceptability range, while hand gestures and the pen fell in the low marginal acceptability range.
Compared with RViz, the XR-based interfaces taken together led to a higher raw workload (35.1 against 25.6, $p = .035$) and frustration ($p = .028$), but not to a different usability ($p = .515$).
Concerning the overall interaction, the participants found it fluent (4.1 out of 5) and felt safe despite a human-sized robot moving around them (1.5 out of 5 on the negated question), while the robot's intentions were rated rather predictable (3.9) and its actions neutrally trusted (3), both with high standard deviations (1.20 and 1.33).


\subsection{Quantitative Results: Task Efficiency}

Table~\ref{tab:device_results} (quantitative results) summarizes the objective task efficiency metrics.
The control method had no significant effect on the task completion time ($p = .601$) nor on the task waiting time ($p = .538$), as both metrics strongly depend on the autonomous navigation of the robot.
Instead, the target location had a significant effect on the completion time ($p = .004$, $\eta_p^2 = .46$), without interaction with the control method ($p = .144$):
the kitchen table, considered the most difficult location, yielded a mean completion time of 79.69~s, longer than the 52.99~s of the living room shelf ($p = .016$) and the 48.89~s of the bed table ($p = .046$).

The effect of the control method on the task selection time did not reach significance ($p = .105$, $\eta_p^2 = .20$):
the pen yielded the shortest mean selection time, 2.44~s, with the smallest standard deviation, and hand gestures the longest, but no pairwise difference was significant ($p \geq .271$).
Neither the target location ($p = .536$) nor its interaction with the control method ($p = .067$) had a significant effect on this metric.
The number of navigation requests did not differ significantly between the four control methods ($p = .064$), but RViz required fewer requests than the XR-based interfaces taken together (4.65 against 9.74, $p = .030$), as the map allowed the participants to directly select the final destination.
The number of requests also depended on the target location ($p = .027$), with more requests for the kitchen table (11.41) than for the living room shelf (6.58, $p < .001$).


\section{Discussion}
\label{sec:discussion}


\subsection{User Experience}

Hypothesis H1, on the user experience of the XR-based interfaces compared to RViz, was not supported:
taken together, the XR-based interfaces led to a higher workload and frustration than RViz without improving the usability, a result driven by hand gestures and the pen, as the motion controllers obtained ratings comparable to those of RViz.
Hypothesis H2, on the user experience of the pen compared to the other XR-based interfaces, was not supported either, as the pen obtained the worst mean usability score and a workload close to that of hand gestures, so that Research Question Q1 could not be answered positively.
However, the user experience of the pen was strongly influenced by a technical limitation discovered during the experiments:
when held still, the pen was detected as inactive and the headset switched to hand tracking, occasionally freezing the selection visuals until manually reset.
The manufacturer of the pen later traced this behavior to the operating system of the headset, which switches from a tracked stylus to hand tracking after a few seconds without stylus motion or button input, an option that can be disabled in its settings\footnote{The authors thank Aidan Kehoe of Logitech for this explanation, also available at \url{https://logitech.github.io/mxink/FAQ.html}.}.
During the experiments, setting the sensitivity of the pen to high brought no noticeable improvement, and the issue was mitigated by asking the participants to gently shake the pen when idle.
As the issue was identified only after several participants had taken part, then reappeared occasionally, and is not captured by the recorded data, its frequency and effect cannot be quantified, so the results of the pen, especially its ratings, should be interpreted with caution.
It never occurred during the practice sessions, and selections with the pen appeared particularly efficient in the trials it did not affect, but whether resolving it would make its usability competitive remains to be verified.
Many participants also struggled to remember the unlabeled selection button of the pen, and several rested their arms when pointing with the pen or hand gestures, consistent with their higher physical demand.

Hand gestures were affected by another technical limitation:
the pinch gesture was often falsely detected at rest, as the resting pose is close to the pinch and the hands naturally point towards the ground, sending unwanted navigation goals near the participant's feet.
The participants noticed it and extended their fingers to avoid false detections, which confirms the concerns raised in previous studies about the pinch gesture~\cite{tadeja_using_2024, kennel-maushart_interacting_2023} and suggests that a selection gesture further away from the natural resting pose of the hand would increase the perceived usability of hand-gesture interfaces.


\subsection{Task Efficiency}

Hypothesis H3, on the task efficiency of the XR-based interfaces compared to RViz, was not supported:
the selection times of the XR-based interfaces taken together did not differ from those of RViz ($p = .586$), and the interaction between control method and target location, on which the part on difficult and occluded targets relies, was not significant.
Hypothesis H4, on the task selection times of the pen compared to the other XR-based interfaces, was not statistically supported either:
the pen produced the shortest mean task selection time with the smallest variability, but, despite a large effect size, this difference did not reach significance with 10 participants.
Moreover, a faster initial selection would not make the overall task more efficient:
the pen required about twice as many navigation requests as RViz (9.54 against 4.61) for similar completion times, so that Research Question Q2 could not be answered positively.
The task completion and waiting times were instead dominated by the autonomous path planning of the robot rather than by the control method:
despite fine-tuned parameters, the robot took tight turns near obstacles and frequently stopped to rescan, frustrating the participants regardless of the interface and, as mentioned several times, probably affecting their questionnaire answers.
The high number of requests with hand gestures also reflects re-selections out of frustration, as observed by the experimenters.
Nevertheless, participants with previous robot experience developed waypoint strategies, forcing the robot along simple paths and reusing almost the same path with all four methods.
This prior experience of some participants, together with the small sample size, limits the generalization of these results to novice users.


\subsection{Observations and Participant Feedback}

No participant was visibly uncomfortable with the robot, and several talked to it, motivating future multimodal control interfaces including natural language.
When using RViz, most participants frequently checked the scene behind them to confirm the robot's actions, suggesting that AR-based interfaces can reduce the mental load by keeping the operator in a single field of view.
During dynamic trials, several participants remained seated yet performed as well as the others, suggesting the interfaces suit users unable to move, whereas one otherwise-seated participant stood up only with hand gestures to reach the kitchen table, again indicating they are less efficient for distant targets.
Finally, several expert participants appreciated the AR-based interfaces and asked for a cancel function, omitted to match RViz but likely to reduce the frustration from false positives.


\section{Conclusion}
\label{sec:conclusion}

This paper compared three XR-based interfaces for semi-autonomous mobile robot navigation in AR, using the XR motion controllers, hand gestures, and a commercial XR pen, with RViz in a user study with 10 participants.
Regarding the integration of the XR pen, the pen yielded the shortest mean selection time with the smallest variability, but without statistical significance and without extending to the overall task efficiency, so that Hypothesis H4 was not statistically supported, and neither was Hypothesis H2, its user experience being held back by a tracking issue, since traced to the automatic switching of the headset to hand tracking.
Regarding the comparative evaluation, the control method significantly affected the perceived workload and usability:
the XR motion controllers obtained the best mean scores, comparable to those of the computer-based baseline, whereas hand gestures led to a significantly higher workload than both, so that Hypotheses H1 and H3 were not supported, and XR-based control remains to be validated without the tracking issue of the pen and with more novice users.

Future work includes reevaluating the XR pen with the automatic switching to hand tracking disabled, investigating selection gestures further from the resting pose of the hand, and adding a cancel function, natural-language interaction, and hands-free inputs based on electroencephalography~\cite{xu_combination_2024} or eye tracking~\cite{tadeja_using_2024}.
By letting operators command a semi-autonomous mobile robot from within the scene, the proposed XR-based interfaces form a building block toward an avatar-symbiotic society~\cite{ishiguro_realisation_2021}.






\balance



\bibliographystyle{IEEEtran}
\bibliography{references}


\end{document}